\documentclass{article}

\usepackage{microtype}
\usepackage{graphicx}
\usepackage{subcaption}
\usepackage{amsmath}
\usepackage{booktabs} 
\usepackage{multirow}
\usepackage{placeins}
 
\usepackage{xcolor} 
\usepackage{tikz}
\usetikzlibrary{fit,calc}

\colorlet{pink}{red!40}
\colorlet{blue}{cyan!60}

\usepackage[compact]{titlesec} 
\titlespacing*{\section}{14pt}{7pt}{4pt}
\titlespacing*{\subsection}{6pt}{3pt}{1pt}
\usepackage{hyperref}

\RequirePackage[colorinlistoftodos,bordercolor=orange,backgroundcolor=orange!20,linecolor=orange,textsize=scriptsize]{todonotes}
\providecommand{\comment}[2]{\todo[inline,caption={}]{\textbf{#1: }#2}}%
\providecommand{\inlinecomment}[3]{%
  {\color{#1}#2: #3}}%
\newcommand\commenter[2]%
{%
  \expandafter\newcommand\csname i#1\endcsname[1]{\inlinecomment{#2}{#1}{##1}}
  \expandafter\newcommand\csname #1\endcsname[1]{\comment{#1}{##1}}
}

\usepackage{enumitem}

\usepackage[accepted]{icml2021}

\icmltitlerunning{On Second-order Optimization Methods for Federated Learning}

\begin{document}

\twocolumn[
\icmltitle{On Second-order Optimization Methods for Federated Learning}



\icmlsetsymbol{equal}{*}

\begin{icmlauthorlist}
\icmlauthor{Sebastian Bischoff}{epfl,tum}
\icmlauthor{Stephan Günnemann}{tum}
\icmlauthor{Martin Jaggi}{epfl}
\icmlauthor{Sebastian U. Stich}{epfl}
\end{icmlauthorlist}

\icmlaffiliation{epfl}{École Polytechnique Fédérale de Lausanne, Lausanne, Switzerland}
\icmlaffiliation{tum}{Technical University of Munich, Munich, Germany}

\icmlcorrespondingauthor{Sebastian Bischoff}{sebastian@salzreute.de}
\icmlcorrespondingauthor{Sebastian Stich}{sebastian.stich@epfl.ch}

\icmlkeywords{Machine Learning, ICML}

\vskip 0.3in
]



\printAffiliationsAndNotice{}  

\newcommand{\GIANTcomrounds}{three }

\begin{abstract}
We consider federated learning (FL), where the training data is distributed across a large number of clients. The standard optimization method in this setting is Federated Averaging (FedAvg), which performs multiple local first-order optimization steps between communication rounds. In this work, we evaluate the performance of several second-order distributed methods with local steps in the FL setting which promise to have favorable convergence properties.\\
We (i) show that FedAvg performs surprisingly well against its second-order competitors when evaluated under fair metrics (equal amount of local computations)---in contrast to the results of previous work. 
Based on our numerical study, we propose (ii) a novel variant that uses second-order local information for updates and a global line search to counteract the resulting local specificity.
\end{abstract}

\newcommand{\func}{f}
\newcommand{\currentstep}{t}
\newcommand{\clients}{\mathcal{S}}
\newcommand{\loss}{l}
\newcommand{\selectedclient}{i}
\newcommand{\weights}{\mathbf{w}}
\newcommand{\data}{\mathbf{x}}
\newcommand{\class}{y}
\newcommand{\numdatapoints}{n}
\newcommand{\update}{\mathbf{u}}
\newcommand{\stepsize}{\mu}
\newcommand{\tgrad}{\mathbf{g}}

\section{Introduction}
\label{intro}
Federated learning (FL) is a new machine learning paradigm where the used data can not be sent to a central server and the individual clients have to participate in the optimization process~\cite{mcmahan2017communication,kairouz2019advances}.
These limitations make classical optimization algorithms not directly applicable to FL, which spurred the interest of the optimization community.
The high iteration cost---due to the need to communicate for each update---requires algorithms that make as much progress as possible per round of communication. 
Second-order methods promise to be a class of algorithms which can do this on a suitable class of functions.

The communication cost in the cross-silo federated learning~\cite{kairouz2019advances} is between the one of classical distributed optimization and cross-device federated learning.
Assuming that clients are reliable and participate in each round, we can formulate a distributed problem of the form:
\begin{equation}
    \func(\weights)= \frac{1}{|\clients|} \sum_{\selectedclient \in \clients} l_\selectedclient(\weights) + \frac{\gamma}{2} \lVert \weights \rVert_2^2 \,,
    \label{eq:classical_distr}
\end{equation}
with regularization as studied in~\citep{wang2018giant} and where all losses $l_\selectedclient(\weights)$ are convex and $\clients$ denoting the set of all clients defining this problem.

The optimization wall-clock time in the cross-device federated learning~\cite{kairouz2019advances} setting is dominated by high latency and low bandwidth connections. Often only a fraction of all (stateless\footnote{Stateless in this setting means that the clients do not have access to any information from previous communication rounds, e.g.\ accumulated gradients, unless communicated by the server.}) clients participate in each round.
This leads to a stochastic approximation of Eq.~\eqref{eq:classical_distr} in the form of
\begin{equation}
    \func_\currentstep(\weights) = \frac{1}{|\clients_\currentstep|} \sum_{\selectedclient \in \clients_\currentstep} l_\selectedclient(\weights) + \frac{\gamma}{2} \lVert \weights \rVert_2^2 \,,
    \label{eq:fed_learn}
\end{equation}
as the set of participating clients $\clients_\currentstep$ changes at each step~$\currentstep$.
The optimization algorithm nevertheless has to optimize problem~\eqref{eq:classical_distr} to find the global optimum.
This setting permits relatively expensive updates as the communication costs usually outweigh the computation costs by far (especially for first-order methods). Such expensive updates can be the ones from second-order methods or even multiple steps of a cheap method.
These local steps before synchronizing the updates make classical convergence rates for a non-distributed setting not directly transferable to this setup.
The changing nature of Eq.~\eqref{eq:fed_learn} becomes problematic when the optimum on a single client is not the optimum of all clients, e.g.\ in the realistic case when data is heterogeneous. Here, a better local solution is possibly not helping to solve the global problem. While it is a challenge for distributed algorithms in general, FedAvg facilitates a balance between the quality of the local update and the overall progress by varying the number of local steps, which is more challenging for second order methods which only do few steps to achieve the same progress.

GIANT~\cite{wang2018giant} is a proposed distributed second order method to optimize Eq.~\eqref{eq:classical_distr} which has better communication complexity than first-order accelerated gradient descent~\citep[Tab.~1]{wang2018giant}.
GIANT calculates an approximation of the Newton update $[\nabla^2 \func(\weights^\currentstep)]^{-1} \nabla \func(\weights)$ by using the global gradient and the local Hessians on each client combined with a global backtracking line search over all clients.
This involves three rounds of communication\footnote{We define sending and receiving of a gradient (or model/state vector) with $\mathcal{O}(d)$ elements as one communication round.} which can be considered quite expensive in comparison with for example local SGD~\cite{Zinkevich2010:local,stich2018local} which only needs one communication round and GIANT does not allow local steps.
LocalNewton~\cite{gupta2021localnewton} uses only the local gradient and local Hessian for its updates combined with a local backtracking line search.
This allows to do multiple local steps before communicating only once for a global update step but with the risk of calculating an update too specific to the client.
When we directly apply both methods to the cross-silo federated learning setting defined by Eq.~\eqref{eq:fed_learn}, GIANT has a disadvantage compared to other methods by its many communication rounds and LocalNewton can find a solution which is good for the current client but does not decrease the objective function overall.

\textbf{Contributions.} Our contributions can be summarized as:
\begin{itemize}[nosep,leftmargin=12pt]
    \item We investigate different new variants of distributed second-order optimizers and derive \textit{LocalNewton with global line search} which is suitable for the federated learning setting, especially with heterogeneous data meaning that each client has a specific data distribution.
    \item We show empirically that FedAvg with multiple local steps is surprisingly effective in the cross-silo and cross-device setting, performing as well as the second-order methods in our experiments.
    \item We argue for a fairer comparison of first and second order methods in distributed optimization by considering the number of gradient evaluations.
\end{itemize}

\section{Related Work}
\label{related}
FedAvg~\cite{mcmahan2017communication} and Local SGD~\cite{stich2018local} use only the first-order gradients for their updates.
The considerable faster computation compared to computing a Hessian is a big advantage over second order methods in settings with fast communication.
Adaptive step size methods like AdaGrad~\cite{duchi2011adaptive}, RMSProp~\cite{tieleman2012lecture} and ADAM~\cite{kingma2014adam}, which MIME~\cite{karimireddy2020mime} transfers to the distributed setting, are trying to improve the convergence properties further.
The step size adaption can also be expressed as a diagonal matrix $D$ and represents in the update $D^{-1} \nabla f(x)$ an alternative preconditioning to the Hessian in $H^{-1} \nabla f(x)$ of Newton's method.
The advantage of local steps is studied in \cite{woodworth2020local,karimireddy2019scaffold,%
woodworth2021min}. \looseness=-1

The proposed second order methods can be divided in methods which use second order information indirectly~\cite{shamir2014communication,li2019feddane,reddi2016aide} and methods which calculate them explicitly~\cite{zhang2015disco,wang2018giant,gupta2021localnewton,zhang2020distributed,crane2019dingo,ghosh2020communication}.
DANE~\cite{shamir2014communication} calculates a mirror descent update on the local function (Eq.~\eqref{eq:local_func}) which is equal to the GIANT update for a quadratic function.
\citet{li2019feddane} propose FedDANE as a version of DANE for federated learning.
They use FedAvg as baseline with 20 local epochs and see no improvement with their proposed method.
AIDE~\cite{reddi2016aide} is an accelerated inexact version of DANE.
Another category are distributed quasi-newton methods like in \citep{agarwal2014reliable}. CoCoA~\cite{smith2018cocoa} and its trust-region extension \cite{duenner2018trust} also perform local steps on a second-order local subproblem, but only address the special case of generalized linear model objectives. \citet{pmlr-v84-karimireddy18a} study inexact updates with global curvature information.

The methods which use the Hessian are calculating it indirectly with the so-called \textit{Hessian-free optimization}~\cite{martens2010deep} approach.
DiSCO~\cite{zhang2015disco} only computes the Hessian-vector-product on the machines and performs the (preconditioned) conjugate gradient method~\cite{hestenes1952methods} on the server which results in one communication round for each conjugate gradient iteration.
GIANT~\cite{wang2018giant} and LocalNewton~\cite{gupta2021localnewton} perform the conjugate gradient method on the clients whereas GIANT uses the global gradient and LocalNewton the local gradient.
\citet{islamov2021distributed} and their FL extension~\cite{safaryan2021fednl} iteratively build an approximation to the global Hessian using a similar amount of communications rounds than GIANT but achieve a better convergence rate.
They use stateful clients for FedNL and their construction works with the assumption that the number of iterations goes to infinity.
One needs to investigate if their experimental results also hold with the few communication rounds we saw in our experiments.
They do not experimentally compare their methods with FedAvg with multiple steps.

\section{Method}
\begin{table*}[t]
\centering
\scalebox{0.95}{
\begin{tabular}{lccccccc}
\toprule 
 & \multicolumn{2}{c}{definition} & & \multicolumn{4}{c}{properties} \\ \cmidrule{2-3}\cmidrule{5-8} 
                            & local & server  & & global & global & local & \#comm. \\
                       & optimization  & update & & gradient & line search & steps & rounds \\
                       \midrule
GIANT~\cite{wang2018giant}  &   Alg.~\ref{alg:GIANT-local}              &  Alg.~\ref{alg:GIANT-linesearch}                              &    & yes             &   yes              & no          & 3                      \\
GIANT with local steps and global line search\textsuperscript{*} & Alg.~\ref{alg:GIANT_local_steps-local} & Alg.~\ref{alg:GIANT-linesearch} & & yes            &    yes              & yes         & 3                      \\
GIANT with local steps and local line search\textsuperscript{*} & Alg.~\ref{alg:GIANT_local_steps_local_linesearch-local} & Alg.~\ref{alg:local_linesearch-update} & & yes             &   no               & yes         & 2                      \\
LocalNewton with global line search\textsuperscript{*} & Alg.~\ref{alg:LocalNewton_global_linesearch-local} & Alg.~\ref{alg:LocalNewton-linesearch}     &  & no & yes & yes & 2 \\
LocalNewton~\cite{gupta2021localnewton}  &  Alg.~\ref{alg:LocalNewton-local}                 &  Alg.~\ref{alg:local_linesearch-update}     & & no              &   no               & yes         & 1  \\ \bottomrule
\end{tabular}
}
\caption{Definition of studied algorithms by used local optimization algorithms and server updates and resulting properties of these algorithms. New methods proposed by us are marked with \textsuperscript{*}. Communication rounds are counted per parameter update on the server.}
\label{tabl:methods}
\end{table*}
\label{method}
\begin{figure}[tb]
\vspace{-2mm}
\begin{algorithm}[H]
    \begin{algorithmic}
    \WHILE{not converged}
        \STATE \hspace{-1ex} $\triangleright$ Select active subset  $\clients_\currentstep \subset \clients$ of clients
        \STATE \hspace{-1ex} $\triangleright$ \textit{Optional:} Compute global gradient on active clients
        \STATE \hspace{3ex} Send parameters $\weights^{t}$ to active clients $\clients_\currentstep$
        \STATE  \hspace{3ex} Compute $\nabla f_\selectedclient(\weights^\currentstep)$ on clients $\clients_\currentstep$ and send to server
        \STATE  \hspace{3ex} $\nabla f_\currentstep(\weights^t) = \frac{1}{|\clients_\currentstep|} \sum_{i\in \clients_\currentstep} f_\selectedclient(\weights^\currentstep)$
        \STATE  \hspace{3ex}  Send $\nabla f_\currentstep(\weights^\currentstep)$ to clients
            \STATE \hspace{-1ex} $\triangleright$ Optimize local functions on active clients $\clients_\currentstep$
            \STATE \hspace{2ex} with Algorithm~\ref{alg:GIANT-local},
            \ref{alg:GIANT_local_steps-local},
            \ref{alg:GIANT_local_steps_local_linesearch-local},
            \ref{alg:LocalNewton_global_linesearch-local} or \ref{alg:LocalNewton-local}
        \STATE \hspace{-1ex} $\triangleright$ Compute update on server
        \STATE \hspace{3ex} by a global backtracking line search (Alg.~\ref{alg:GIANT-linesearch}),
        \STATE \hspace{3ex} or by averaging weights (Alg.~\ref{alg:local_linesearch-update}),
        \STATE \hspace{3ex} or by a global line search (Alg.~\ref{alg:LocalNewton-linesearch}).
    \ENDWHILE
    \end{algorithmic}
\caption{Blueprint of all methods from Table~\ref{tabl:methods}}
\label{alg:blueprint}
\end{algorithm}
\end{figure}
An optimization algorithm for the federated learning setting defined in Eq.~\eqref{eq:fed_learn} only has access to the subset $\clients_\currentstep$ of all clients $\clients$ in each timestep $\currentstep$ and each client $\selectedclient$ can only optimize its local objective 
\begin{equation}
    f_\selectedclient(\weights)=l_\selectedclient(\weights)+ \frac{\gamma}{2} \lVert \weights \rVert_2^2
    \label{eq:local_func}
\end{equation} individually.
The function $l_\selectedclient(\weights)$ can be any convex function.
The regularization term $\frac{\gamma}{2} \lVert \weights \rVert_2^2$ makes each local function $f_\selectedclient(\weights)$ strongly convex which implies a positive definite Hessian.
The function over all clients is then defined by Eq.~\eqref{eq:classical_distr} whereby the optimization algorithm only has access to the stochastic approximation Eq.~\eqref{eq:fed_learn} in each step.

Implementing Newton's method naively by using the global gradient and sending the local Hessians $H_{\selectedclient,\currentstep}=\nabla^2 f_\selectedclient(\weights^\currentstep)$ to the server that would then calculate the global Hessian by $H_\currentstep=\frac{1}{|\clients_\currentstep|} \sum_{\selectedclient\in\clients_\currentstep} H_{\selectedclient,\currentstep}$ and invert it for the update
\begin{equation}\textstyle
    \weights^{\currentstep+1}=\weights^{\currentstep} - \stepsize_t \left(\frac{1}{|\clients_\currentstep|} \sum_{\selectedclient\in\clients_\currentstep} H_{\selectedclient,\currentstep} \right)^{-1} \nabla \func_\currentstep(\weights^{\currentstep})
    \label{eq:dist_newton}
\end{equation}
leads to a communication and space complexity of $\mathcal{O}(|\clients_\currentstep| \cdot d^2)$ which is already prohibitive for a moderate-sized dimensionality $d$.

Instead, we can use the method of \citet{pearlmutter1994fast} to calculate Hessian-vector-products in conjunction with solving $H_\selectedclient \update_\selectedclient = \nabla f_\selectedclient(\weights)$ for $\update_\selectedclient$ with the conjugate gradient method~\cite{hestenes1952methods} which gives us the update $\update_\selectedclient = H_\selectedclient^{-1} \nabla f_\selectedclient(\weights)$ without having to form the Hessian nor invert it explicitly. This involves $\mathcal{O}(n_i q)$ gradient evaluations where $q$ is the number of CG iterations needed and $n_\selectedclient$ is the number of samples on client $\selectedclient$.
We then only have to send the update with size $\mathcal{O}(d)$ to the server which averages them and calculates the new weights as
\begin{equation} \hspace{-1ex}
\resizebox{.92\linewidth}{!}{$
\displaystyle 
    \weights^{\currentstep+1}
    =\weights^{\currentstep}- \frac{\stepsize_t}{|\clients_{\currentstep}|} \sum_{\selectedclient\in\clients_{\currentstep}} \update_{\selectedclient}
    = \weights^{\currentstep}- \frac{\stepsize_t}{|\clients_{\currentstep}|} \sum_{\selectedclient\in\clients_{\currentstep}} H_{\selectedclient{},\currentstep{}}^{-1} \nabla f_{\currentstep}(\weights)$}
    \label{eq:cg_update}
\end{equation}
using a backtracking line search to find $\stepsize_t$.
One can see that the update in Eq.~\eqref{eq:cg_update} is not the same as in Eq.~\eqref{eq:dist_newton} as the former first inverts the Hessians and then averages them instead of the other way round in the correct update. \citet{derezinski2019distributed} show that the used update in Eq.~\eqref{eq:cg_update} is a biased estimate where using more machines to calulcate the update does not result in a better estimate (starting at roughly 100 machines in their experiments).
An open question is how to use their proposed \textit{determinantal averaging} to remedy this problem without having access to an explicit Hessian.

The convergence properties of the conjugate gradient method solving $H_{\selectedclient,\currentstep} \update_\selectedclient = \nabla f(\weights)$ is very important to the overall performance of all discussed second order methods.
The time complexity is $\mathcal{O}(n_i dq)$~\cite{wang2018giant} where $q$ is the number of CG iterations and $d$ is the dimensionality of the problem.
Each evaluation of $Hv$ with an arbitrary vector $v$ takes as much time as one gradient computation~\cite{pearlmutter1994fast}.
A fair comparison between first- and second-order methods therefore uses as many local steps for FedAvg as the second order methods need to iterate the CG method.
GIANT treats maximal iteration for CG additionally as hyperparameter.

A naively implemented backtracking line search needs one communication round for each line search iteration as it potentially decreases the step size continuously. \citet{wang2018giant} propose to use a fixed set of step sizes for which the losses are calculated and then sent to the server in one communication round (See Alg.~\ref{alg:backtrack-linesearch} for details).

Each of the possible algorithms (See Table~\ref{tabl:methods} and Alg.~\ref{alg:blueprint}) can then either use the global gradient $\nabla \func_\currentstep(\weights)=\frac{1}{|\clients_\currentstep|} \sum_{i\in \clients_\currentstep} f_\selectedclient(\weights)$ or only the local gradient $\nabla f_\selectedclient(\weights)$, either use a global line search over Eq.~\eqref{eq:fed_learn} or a local line search only over Eq.~\eqref{eq:local_func} and all of our proposed variants use local steps.
The global gradient and global line search each need one more communication round than their local counterparts but lead to the inclusion of global information in the update process.
\textit{GIANT with local steps and global line search} and \textit{LocalNewton with global line search} introduce an additional step size parameter for the local steps.
The local steps are performed with this additional step size parameter and the global line search is performed over the resulting update from multiple local steps ($u_i = \weights_l^{t}-\weights_0^{t}$).
We select a new active subset of clients for the global line search of \textit{LocalNewton with global line search} similar to the sampling of a new minibatch proposed for the adaption of the Armijo line-search~\cite{armijo1966minimization} in a stochastic setting~\cite{Vaswani2019}.
\textit{LocalNewton with global line search} does not use the global gradient for the update calculation and we would need to calculate it only for the backtracking line search.
This could be done in parallel to the update computations and is therefore possibly not as expensive as for methods needing it for the update.
Instead, we choose the step size using $\arg\min_{\gamma \in \gamma_1, \ldots, \gamma_l} \sum_{i\in \clients_t} f_i(w+\gamma_i u)$ which can potentially choose a bigger step size than original backtracking line search would but the backtracking line search variant of \citet{wang2018giant} with a fixed set of step sizes has the same problem.
Both variants of \textit{GIANT with local steps} can use the real global gradient only in the first local step. Afterwards, they can update the global gradient only by their local gradient with $\tgrad_{j+1} = \tgrad_j - \frac{1}{|\clients_\currentstep|} \nabla f_\selectedclient(\weights_j^\currentstep) + \frac{1}{|\clients_\currentstep|} \nabla f_\selectedclient(\weights_{j+1}^\currentstep)$ as calculating the global gradient at the new parameters $\weights_\selectedclient^{\currentstep+1}$ would need at least one additional communication round.\looseness=-1

\begin{figure*}[t]
     \begin{subfigure}{0.333\textwidth}
        \centering
        \includegraphics[width=\linewidth]{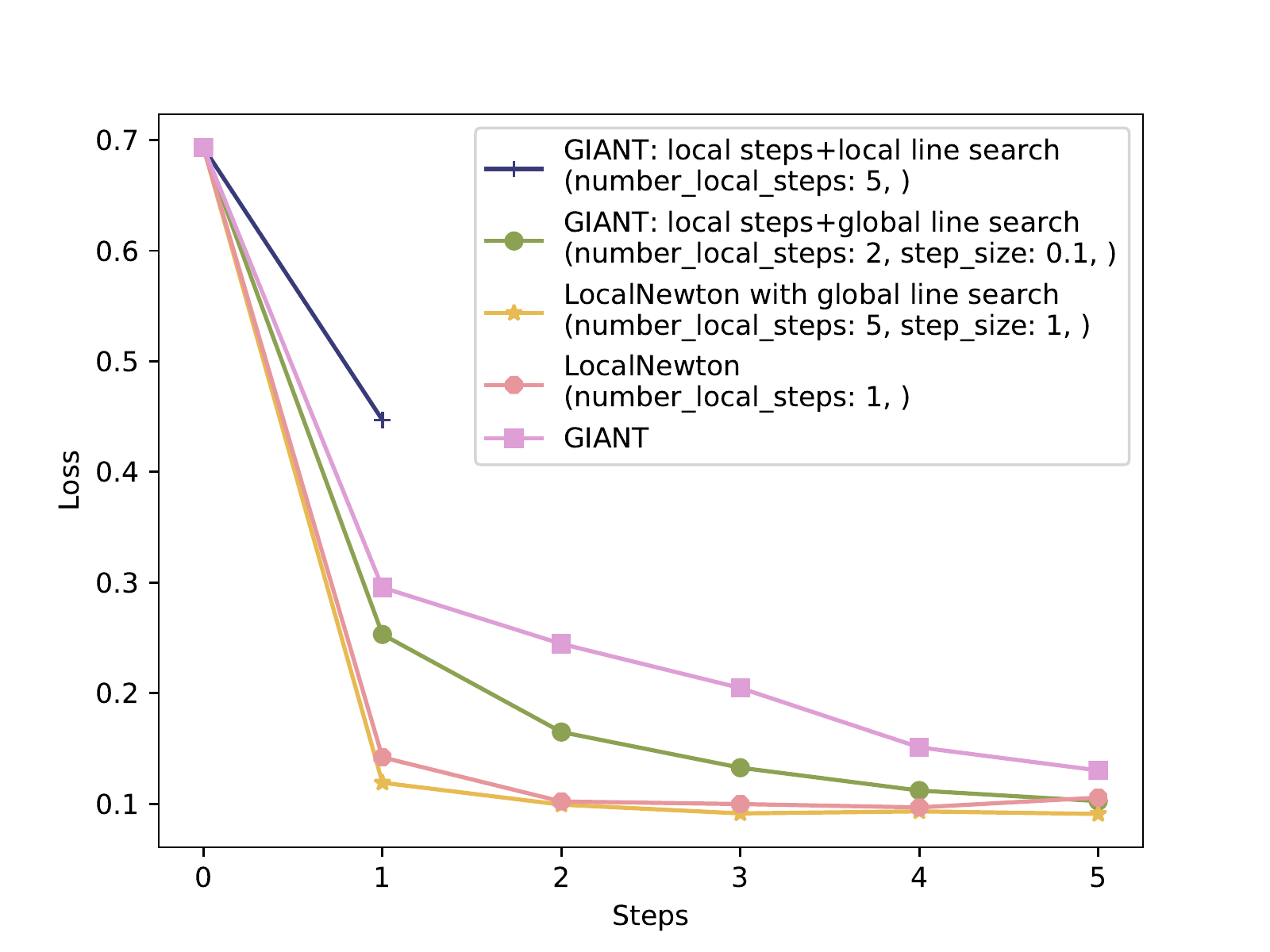}
        \caption{w8a}
        \label{fig:w8a-second_order}
    \end{subfigure}%
    \hspace{0.1em}%
    \begin{subfigure}{0.333\textwidth}
        \centering
        \includegraphics[width=\linewidth]{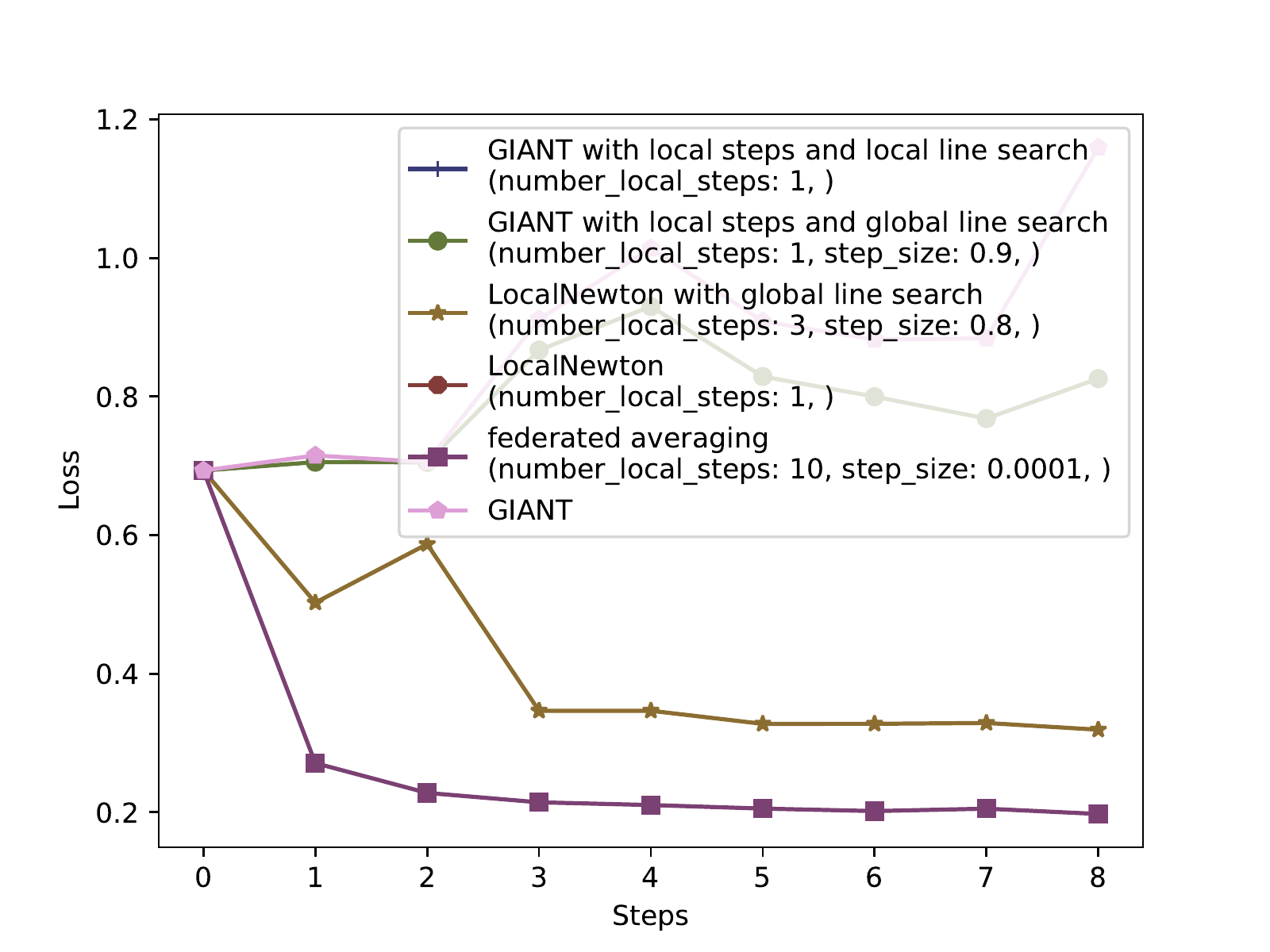}
        \caption{Synthetic heterogenous data}
        \label{fig:toy-noniid}
    \end{subfigure}%
    \hspace{0.1em}%
    \begin{subfigure}{0.333\textwidth}
        \centering
        \includegraphics[width=\linewidth]{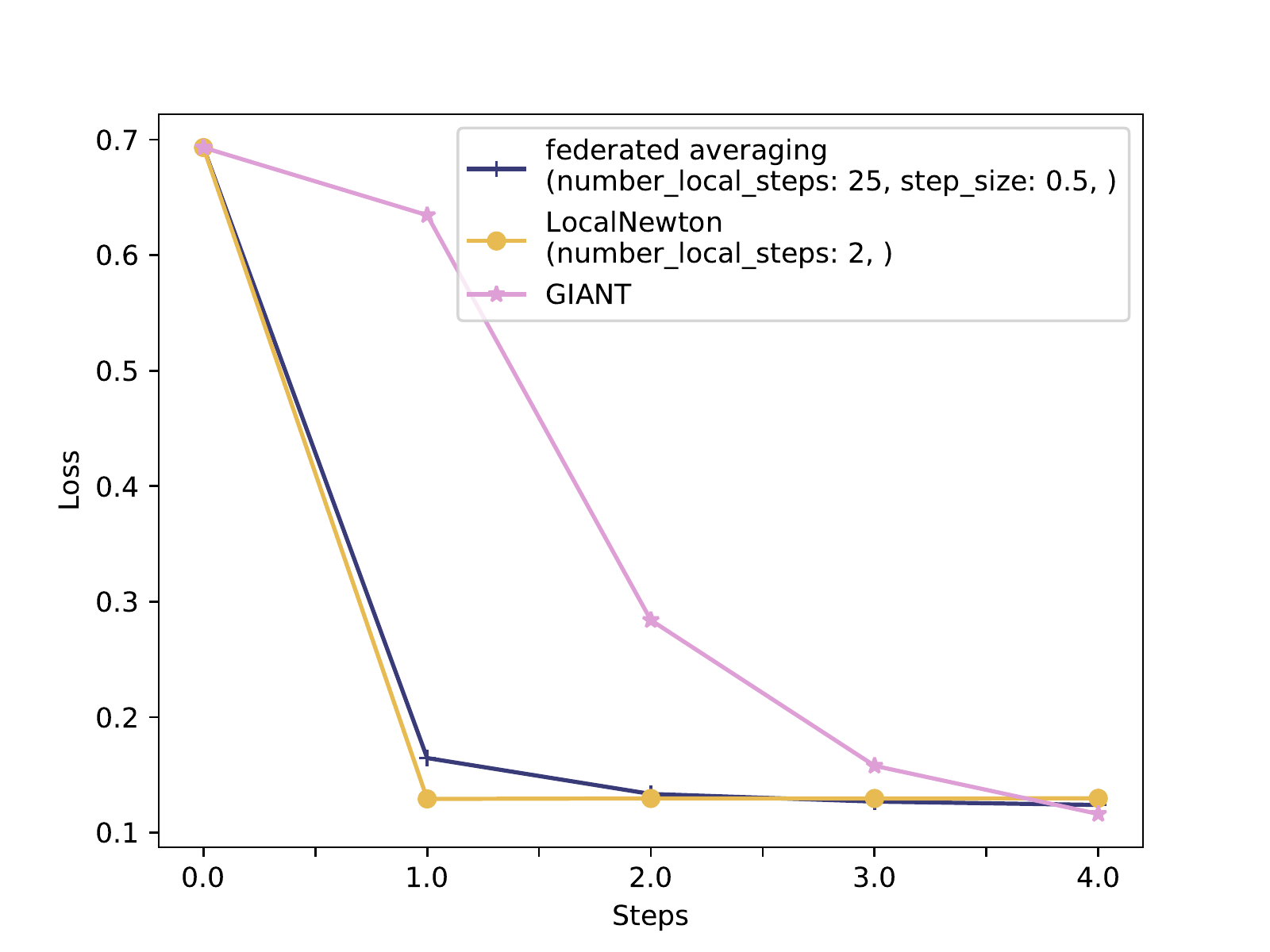}
        \caption{w8a}
        \label{fig:dummy-allclients}
    \end{subfigure}
    \caption{(a) All second order methods on w8a in cross-device setting. LocalNewton variants working best. (b) Second order methods on synthethic non-i.i.d. dataset in cross-device setting. Only LocalNewton with global line search is able to minimize loss. (c) Fair Comparison between Local SGD, GIANT and LocalNewton in cross-silo setting. Local SGD performing as well as second order methods.}
\end{figure*}
\section{Experiments}
\label{experiments}
We perform $\ell_2$-regularized logistic regression over all clients. The optimizers only have access to the subset $\clients_\currentstep$ of all clients in each step $\currentstep$.
The loss on client $\selectedclient$ is therefore \vspace{-0.5mm}
\begin{equation*}
    \loss_\selectedclient(\weights) = -\frac{1}{n} \sum_{j=1}^n \class_j \log(p_j) + (1-\class_j) \log(1-p_j)
\end{equation*} \par \vspace{-3mm}
with $p_j= \frac{1}{1+\exp(\data_j \weights_j)}$. 
A x-axis titled \textit{step} means an update step on the server of the form $\weights^{t+1}=\weights^{t}+\update$ and a x-axis titled communication round means that the server sends $\mathcal{O}(d)$ information to the clients and can receive $\mathcal{O}(d)$ information from each client.
We use grid search to optimize the number of local steps and the step size for the local steps for the methods with global line search and federated averaging.
\citet{gupta2021localnewton} execute local SGD as baseline for one epoch with batch size one which would be $n_\selectedclient$ steps with one gradient evaluation per step and therefore $\mathcal{O}(n_\selectedclient)$ gradient evaluations.
They perform up to $L=3$ local steps for LocalNewton on all $n_i$ local data points resulting in $\mathcal{O}(L q n_i)$, e.g. $3 \times 100 n_i$\footnote{\citet{gupta2021localnewton} do not specify the maximal number of CG iterations but \citet[Figure~7]{wang2017giantArxiv} use maximal 100 CG iterations in one of the experiments.}, gradient evaluations.
\citet{wang2018giant} compare GIANT only with accelerated GD as first order method which only uses $\mathcal{O}(n_\selectedclient)$ gradient evaluations.
The baselines in \citep{wang2018giant} and \citep{gupta2021localnewton} are therefore disadvantaged which we could reproduce in our experiments. FedAvg with $l$ local steps and therefore $\mathcal{O}(l n_\selectedclient)$ gradient evaluations performs significantly better than FedAvg with only one step.
Figure~\ref{fig:dummy-allclients} shows that Local SGD is competitive with second order methods on w8a and Figure~\ref{fig:toy-noniid} shows that FedAvg is better than all tested second-order methods on heterogeneous data.

\textbf{Data}
\label{synth}
We use the dataset w8a from LibSVM~\cite{chang2011libsvm} which we distribute to 50 clients. This would result in 1000 data points per client which can nearly be solved in one step by most methods and is therefore uninteresting for our experiments. Sampling only 10\% of the data differentiates the methods better.
As an additional dataset, we generate data for $\class=0$ with $\data\sim \mathcal{N}(\mu_0+b_\selectedclient, \Sigma_{\selectedclient,0})$ and for $\class=1$ with $\data\sim \mathcal{N}(\mu_1+b_\selectedclient,, \Sigma_{\selectedclient,1})$ where $\Sigma_{\selectedclient,j}=A_{\selectedclient,j}^\top A_{\selectedclient,j}$ with $A_{\selectedclient,j}=\mathcal{U}(0,1)^{d\times d}$ and $b_i=\mathcal{U}(-100,100)^{d}$.
With this setup, we can test the performance of the different methods on i.i.d data by using $b_\selectedclient=0$ and $\forall i,k,  A_{\selectedclient,j}=A_{k,j}$ and on non-i.i.d. data.
In the experiments, $\gamma$ is choosen as $\frac{1}{\numdatapoints}$ with $\numdatapoints=1000$ generated data points. 

\textbf{Results}
Fig.~\ref{fig:w8a-second_order} and Fig.~\ref{fig:dummy-GIANT} show the advantage of adding local steps to GIANT.
The local steps allow the method to make more progress in one communication round and allow to choose multiple steps with a smaller stepsize when a single step with a higher step size would be too noisy.
The difference between Local Newton and \textit{Local Newton with global line search} on an i.i.d. setting like in Fig.~\ref{fig:w8a-second_order} is minimal but the global line search definitely helps in a non i.i.d. setting like in Fig.~\ref{fig:toy-noniid}.
Using a local line search for GIANT and therefore saving one communication round is not working as can be seen in Fig.~\ref{fig:w8a-second_order} and Fig.~\ref{fig:dummy-GIANT}.
This method fails in nearly all experiments.
Table~\ref{tabl:methods} raises the question if one should invest the second communication round into a global gradient (\textit{GIANT with local steps and local line search}) or a global line search (\textit{LocalNewton with global line search}) and our experiments indicate that \textit{LocalNewton with global line search} is advantageous (Fig.~\ref{fig:w8a-second_order} and \ref{fig:dummy-investment}).
Figure~\ref{fig:toy-noniid} shows that the second order methods except for \textit{LocalNewton with global line search} struggle with the non-i.i.d setup with client-specific means.
The GIANT variants with global line search are choosing steps which do not improve the overall loss which is not prevented by using another set of clients for the global line search (not shown here). The two methods with only local line search already choose a too specific first update which results in them diverging.
\textit{LocalNewton with global line search} shows among the second-order methods the best performance overall considering the number of used communication rounds.
Figure~\ref{fig:fair-fed} shows the competitiveness of Federated Averaging also in the cross-device setting when all methods have the same gradient evaluation budget as discussed in Section~\ref{method}.
An interesting empirical observation is that the conjugate gradient methods needs an increasing number of iterations after each update step to converge to a given tolerance.
This makes a fair comparison with first-order methods more difficult. 
\section{Conclusion}
\label{conclusion}
Our work proposed new second-order methods for federated learning, showed that the second-order methods exhibit very different characteristics on i.i.d.\ and heterogenous data, showed surprisingly good results for Local SGD/FedAvg in the cross-silo and cross-device setting and suggested a fairer comparison between first- and second-order methods in distributed optimization.
An interesting question raised is if one can characterize federated learning problems were second-order methods are of advantage.

\FloatBarrier
\clearpage
{\small
\newpage
\bibliographystyle{icml2021}
\bibliography{example_paper}
}

\clearpage
\appendix
\begin{figure}[tb]
\vspace{-2mm}
\begin{algorithm}[H]
    \begin{algorithmic}[h!]
            \STATE Solve $H_{\selectedclient,\currentstep} \update_\selectedclient = \nabla f_\currentstep(\weights^\currentstep)$ for $\update_\selectedclient$ using CG-method
            \STATE Send $\update_\selectedclient$ to server
    \end{algorithmic}
\caption{Local optimization for GIANT}
\label{alg:GIANT-local}
\end{algorithm}
\vspace{-4mm}
\begin{algorithm}[H]
    \begin{algorithmic}
        \STATE $\weights_0^{t} = \weights^{t}$
        \STATE $\tgrad_0 = \nabla f_t(\weights^t)$
        \FOR{$j$ from $0$ to number of local steps $l$}
            \STATE Solve $H_{\selectedclient,j} \update_j = \tgrad_j$  for $\update_j$ using CG-method
            \STATE $\weights_{j+1}^{t} = \weights_{j}^{t} - \gamma \update_j$
            \STATE $\tgrad_{j+1} = \tgrad_j - \frac{1}{|k_t|} \nabla f_\selectedclient(\weights_j^t) + \frac{1}{|k_t|} \nabla f_\selectedclient(\weights_{j+1}^t)$
        \ENDFOR
        \STATE Send $\update_\selectedclient = \weights_l^{t}-\weights_0^{t}$ to server
    \end{algorithmic}
\caption{Local optimization for \textit{GIANT with local steps and global line search} on client $\selectedclient$ with step size $\gamma$}
\label{alg:GIANT_local_steps-local}
\end{algorithm}
\vspace{-4mm}
\begin{algorithm}[H]
    \begin{algorithmic}[h!]
        \STATE $\weights_0^{t} = \weights^{t}$
        \STATE $\tgrad_0 = \nabla f_t(\weights^t)$
        \FOR{$j$ from $0$ to number of local steps $l$}
            \STATE Solve $H_\selectedclient \update_\selectedclient = \tgrad_j$ for $\update_\selectedclient$ using CG-method
            \STATE Choose $\gamma_j$ with local backtracking line search
            \STATE $\weights_{j+1}^{t} = \weights_{j}^{t} - \gamma_j \update_\selectedclient$
            \STATE $\tgrad_{j+1} = \tgrad_j - \frac{1}{|k_t|} \nabla f_\selectedclient(\weights_j^t) + \frac{1}{|k_t|} \nabla f_\selectedclient(\weights_{j+1}^t)$
        \ENDFOR
        \STATE Send $\weights_{l}^{t}$ to server
    \end{algorithmic}
\caption{Local optimization for GIANT with local steps and local line search on client $\selectedclient$}
\label{alg:GIANT_local_steps_local_linesearch-local}
\end{algorithm}
\vspace{-4mm}
\begin{algorithm}[H]
    \begin{algorithmic}
        \FOR{$j$ from $0$ to number of local steps $l$}
            \STATE Solve $H_{\selectedclient,\currentstep} \update_\selectedclient = \nabla f_\selectedclient(\weights^\currentstep)$ for $\update_\selectedclient$ using CG-method
            \STATE $\weights_{j+1}^{t} = \weights_{j}^{t} - \gamma_j \update_\selectedclient$
        \ENDFOR
        \STATE Send $u_i = \weights_l^{t}-\weights_0^{t}$ to server
    \end{algorithmic}
\caption{Local optimization for \textit{LocalNewton with global line search} with step size $\gamma_j$}
\label{alg:LocalNewton_global_linesearch-local}
\end{algorithm}%
\vspace{-4mm}
\begin{algorithm}[H]
    \begin{algorithmic}[h!]
        \STATE $w_0^\currentstep = w^\currentstep$
        \FOR{$j$ from $0$ to number of local steps $l$}
            \STATE Solve $H_{\selectedclient,\currentstep} \update_\selectedclient = \nabla f_\selectedclient(\weights_j^\currentstep)$ for $\update_\selectedclient$ using CG-method
            \STATE Choose $\gamma_j$ with local backtracking line search
            \STATE $\weights_{j+1}^{t} = \weights_{j}^{t} - \gamma_j \update_\selectedclient$
        \ENDFOR
        \STATE Send $\weights_{l}^{t}$ to server
    \end{algorithmic}
\caption{Local optimization for LocalNewton}
\label{alg:LocalNewton-local}
\end{algorithm}
\end{figure}
\begin{figure}
\vspace{-4mm}
\begin{algorithm}[H]
    \begin{algorithmic}
            \STATE $\update
        = \frac{1}{|\clients_\currentstep|} \sum_{i\in \clients_\currentstep} \update_\selectedclient$ 
        \STATE Send $\update$ to clients
        \FOR{clients $\clients_\currentstep^{'}$ in parallel}
            \FOR{predefined step sizes $\stepsize_1, \ldots, \stepsize_l$}
                \STATE Compute $\func_i(\weights^{t}-\stepsize_m u)$
            \ENDFOR
            \STATE Send $\func_i(\weights^{t}-\stepsize_1 u), \ldots, \func_i(\weights^{t}-\stepsize_l u)$ and $\func_i(\weights^{t})$ to server
        \ENDFOR
        \STATE Find optimal stepsize $\stepsize$ with global backtracking line search (Alg.~\ref{alg:backtrack-linesearch})
        \STATE $\weights^{t+1} = \weights^{t} - \stepsize \update$
    \end{algorithmic}
\caption{Line search procedure for GIANT versions with global line search}
\label{alg:GIANT-linesearch}
\end{algorithm}
\vspace{-4mm}
\begin{algorithm}[H]
    \begin{algorithmic}
        \STATE $w^{t+1} = \frac{1}{|k_t|} \sum_{k\in k_t} w_k^{t+1}$
    \end{algorithmic}
\caption{Update on server for methods with local line search}
\label{alg:local_linesearch-update}
\end{algorithm}
\vspace{-4mm}
\begin{algorithm}[H]
    \begin{algorithmic}
        \STATE $\update
        = \frac{1}{|\clients_\currentstep|} \sum_{i\in \clients_\currentstep} \update_\selectedclient$
        \STATE Select new active subset  $\clients_\currentstep^{'} \subset \clients$ of clients
        \STATE Send $\update$ to clients
        \FOR{clients $\clients_\currentstep^{'}$ in parallel}
            \FOR{predefined step sizes $\stepsize_1, \ldots, \stepsize_l$}
                \STATE Compute $\func_i(\weights^{t}-\stepsize_m u)$
            \ENDFOR
            \STATE Send $\func_i(\weights^{t}-\stepsize_1 u), \ldots, \func_i(\weights^{t}-\stepsize_l u)$ to server
        \ENDFOR
        \STATE $\stepsize = \arg\min_{\gamma \in \gamma_1, \ldots, \gamma_l} \sum_{i\in \clients_\currentstep} f_i(w-\gamma_i u)$
        \STATE $\weights^{t+1} = \weights^{t} - \stepsize \update$ 
    \end{algorithmic}
    \caption{Line search procedure for \textit{LocalNewton with global line search}}
\label{alg:LocalNewton-linesearch}
\end{algorithm}
\vspace{-4mm}
\begin{algorithm}[H]
    \begin{algorithmic}
        \FOR{$\stepsize_i$ in $\stepsize_1,\ldots,\stepsize_l$}
            \IF{$ \func_\currentstep(\weights^\currentstep + \stepsize_i \update^\currentstep) \leq \func_\currentstep(\weights^\currentstep) - \stepsize_i c \langle \update^\currentstep, \nabla_\currentstep (\weights^\currentstep) \rangle$}
                \STATE \RETURN $\stepsize_i$
            \ENDIF
        \ENDFOR
        \STATE \RETURN $\stepsize_l$
    \end{algorithmic}
\caption{Line search used for methods with global backtracking line search}
\label{alg:backtrack-linesearch}
\end{algorithm}
\end{figure}

\section{Implementation Details}
\label{implementation}
We tuned the number of local steps and step size for \textit{GIANT with local steps and global line search}, the number of local steps for \textit{GIANT with local steps and local line search}, the number of local steps and step size for \textit{LocalNewton with global line search}, number of local steps for \textit{LocalNewton} and number of local steps and step size for \textit{FedAvg}.
The conjugate gradient method is limited to 250 iterations and initialized with a random initial point.
The second-order methods were for (number of local steps, step size) optimized over $\{0.1, 0.5, 0.6, 0.7, 0.8, 0.9, 1\} \times \{1, 2, 3, 5, 10\}$ and FedAvg over $\{10^{-5}, 10^{-4}, 10^{-3}, 10^{-2}, 0.1, 0.5, 0.9, 1\} \times \{1, 10, 25, 50, 100\}$.
After running all parameter pairs, we selected the one which had the smallest loss after the last step.

\begin{figure*}
     \begin{subfigure}{0.333\textwidth}
        \centering
        \includegraphics[width=\linewidth]{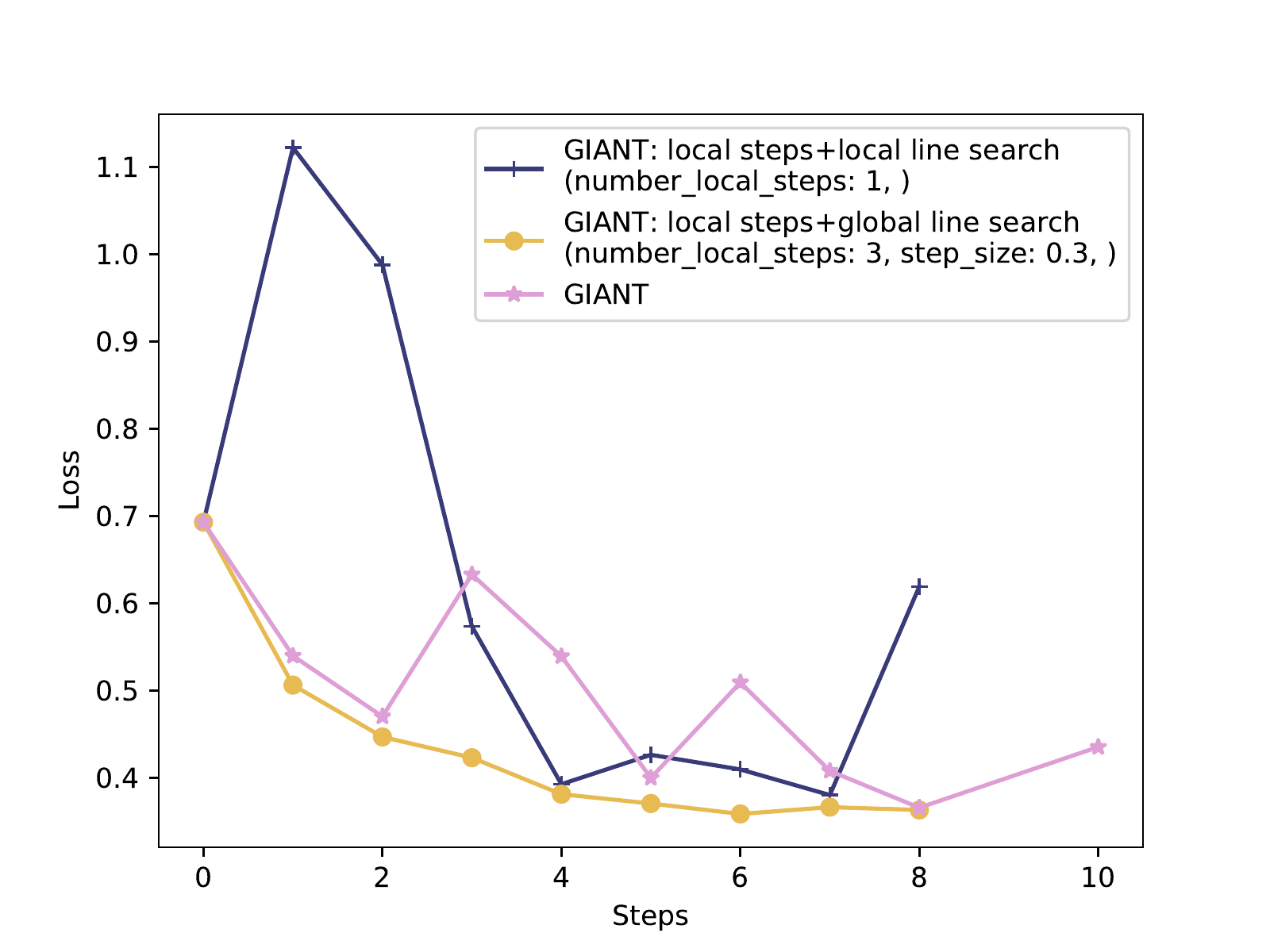}
        \caption{}
        \label{fig:dummy-GIANT}
    \end{subfigure}%
    \hspace{0.1em}%
    \begin{subfigure}{0.333\textwidth}
        \centering
        \includegraphics[width=\linewidth]{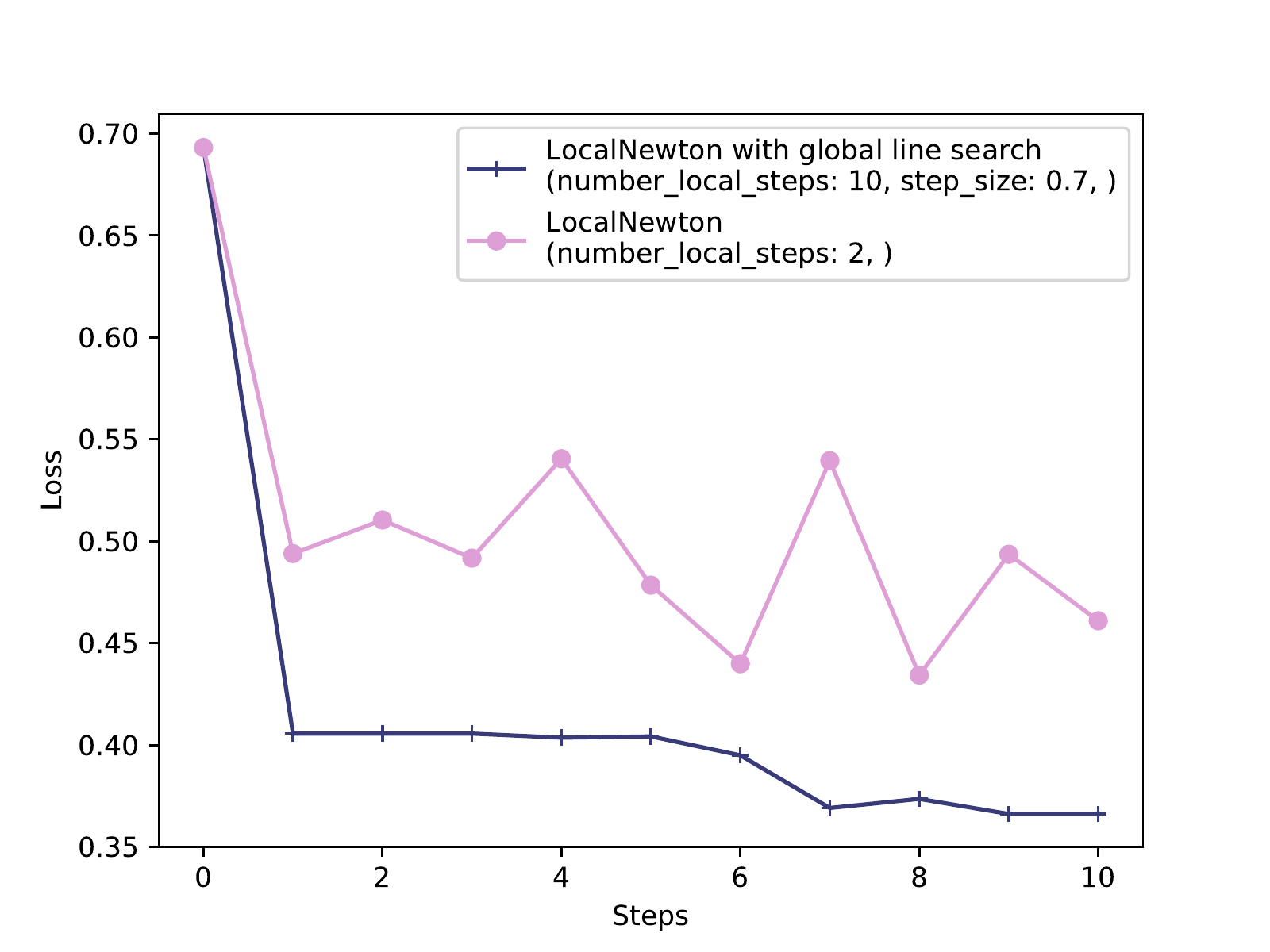}
        \caption{}
        \label{fig:dummy-LocalNewton}
    \end{subfigure}%
    \hspace{0.1em}%
    \begin{subfigure}{0.333\textwidth}
        \centering
        \includegraphics[width=\linewidth]{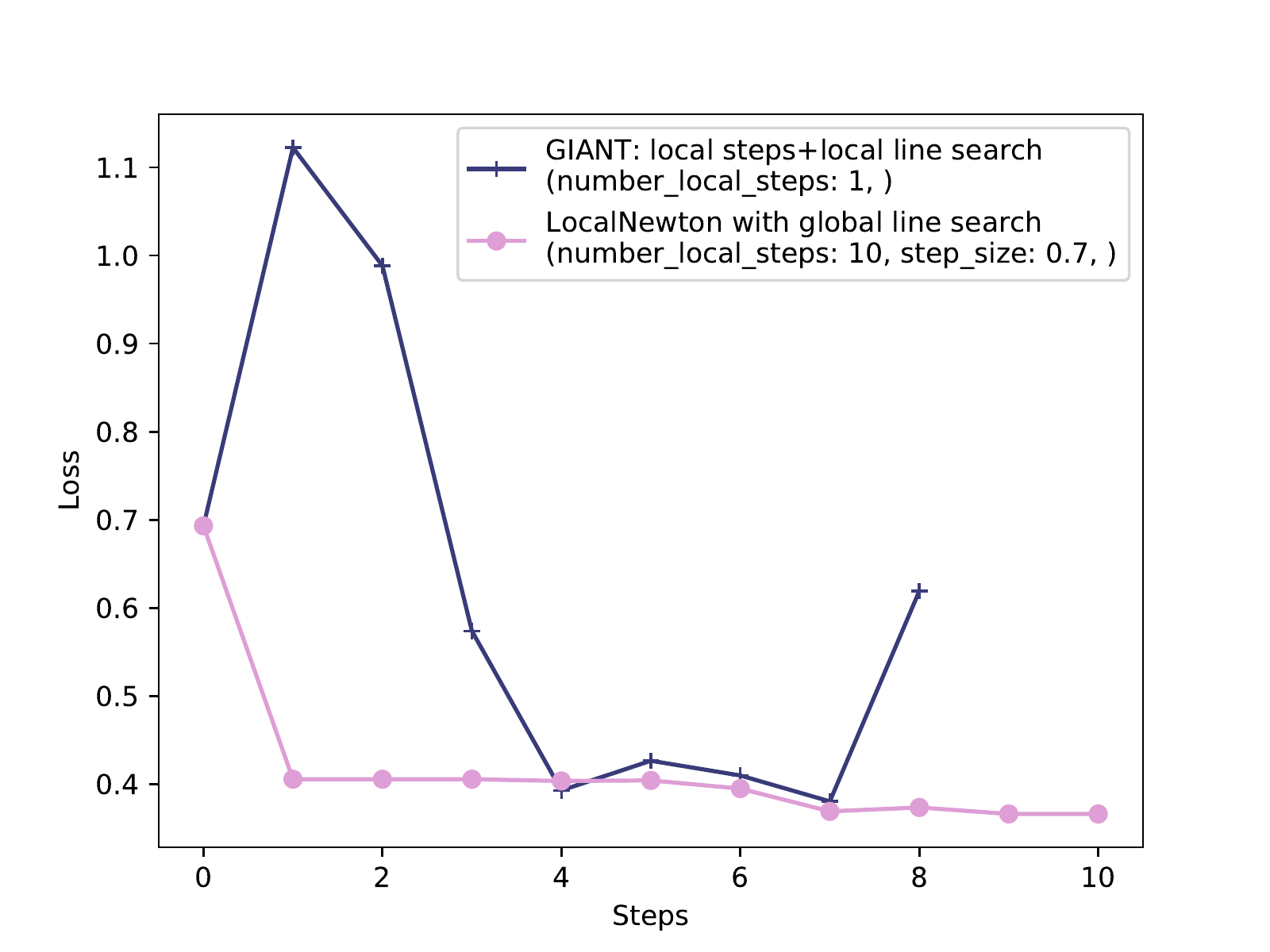}
        \caption{}
         \label{fig:dummy-investment}
    \end{subfigure}
    
    \begin{subfigure}{0.333\textwidth}
        \centering
        \includegraphics[width=\linewidth]{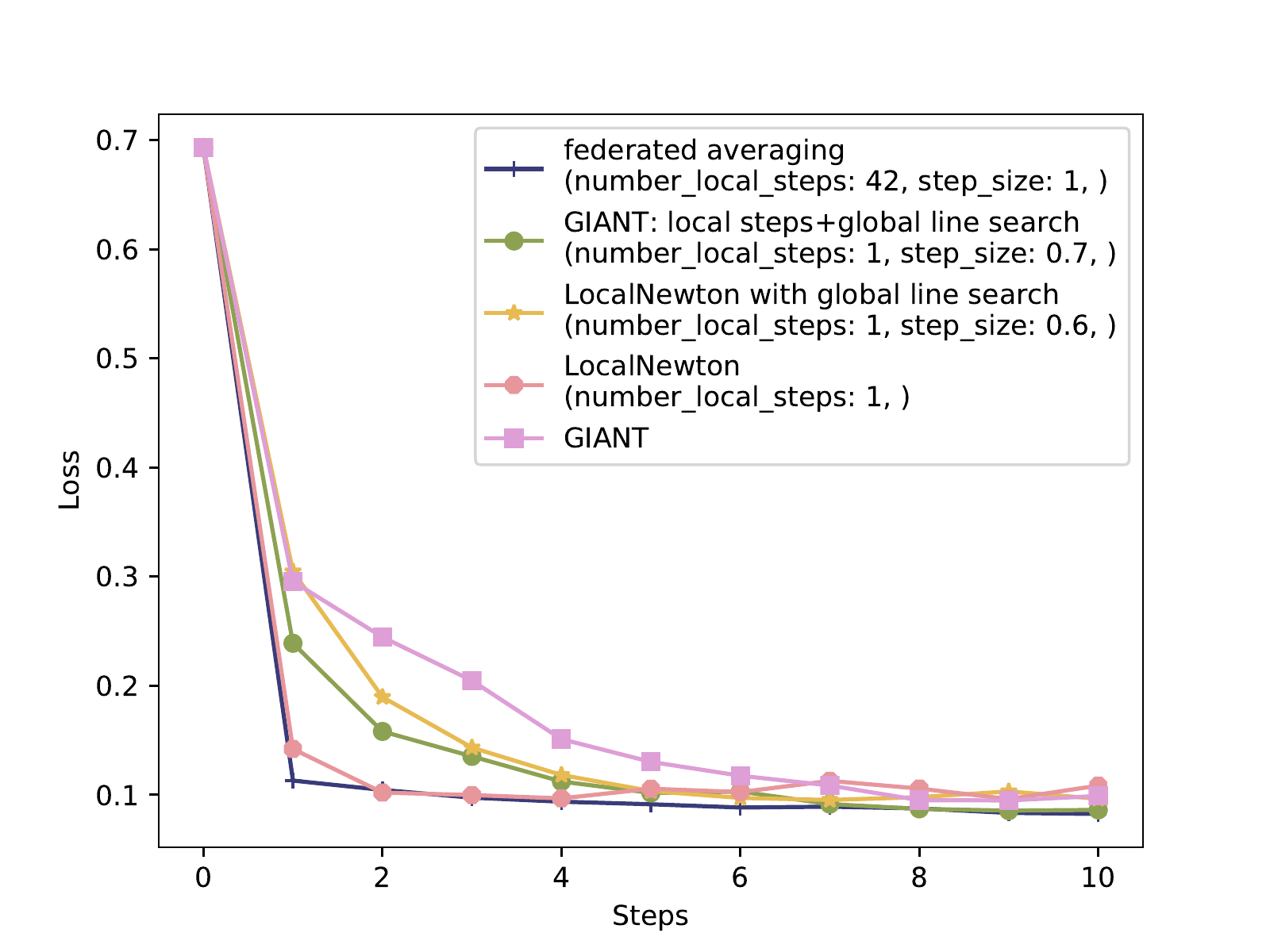}
        \caption{}
        \label{fig:fair-fed}
     \end{subfigure}%
    \hspace{0.1em}%
    \begin{subfigure}{0.333\textwidth}
        \centering
        \includegraphics[width=\linewidth]{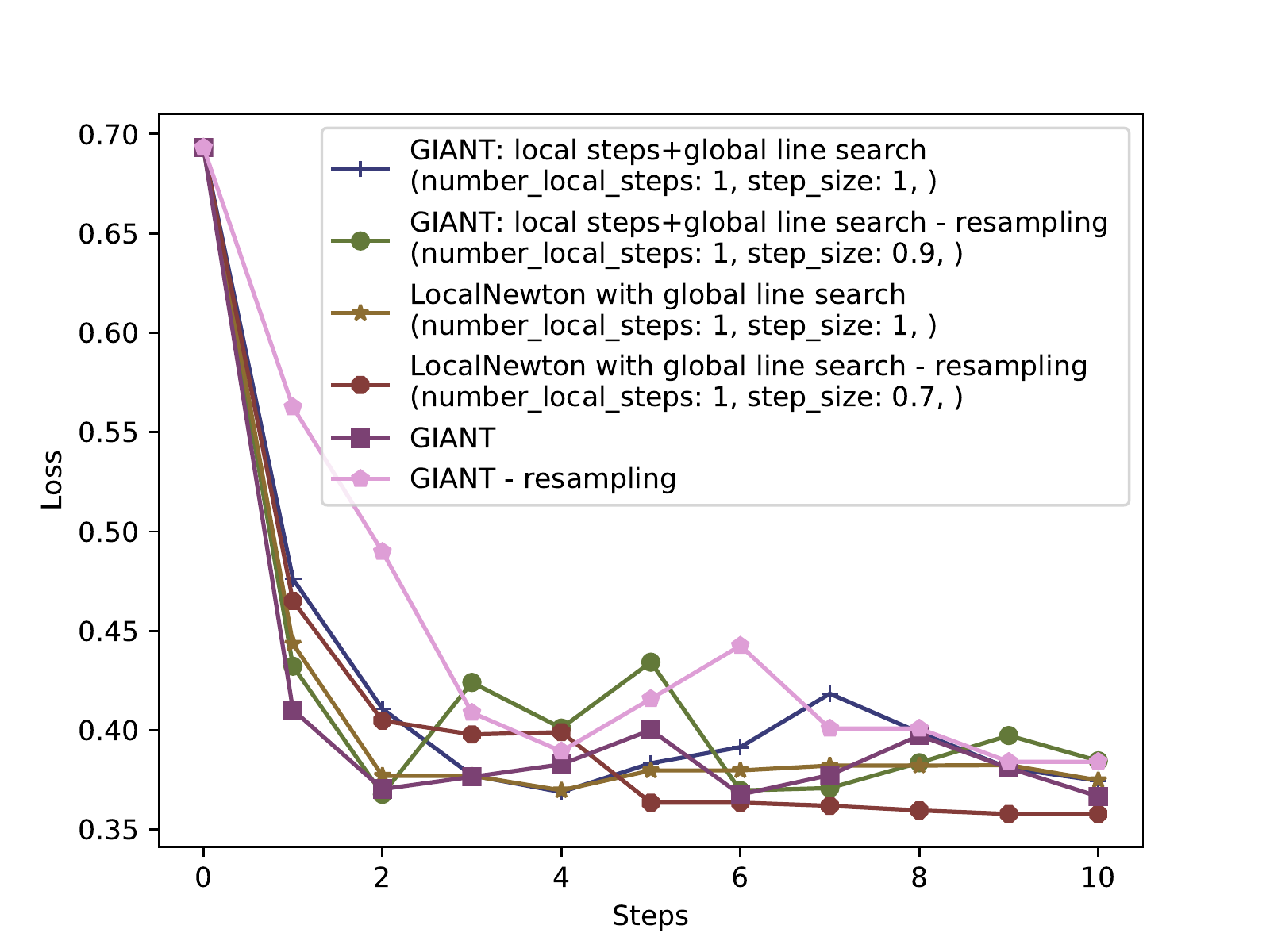}
         \caption{}
     \end{subfigure}%
    \hspace{0.1em}%
    \begin{subfigure}{0.333\textwidth}
        \centering
        \includegraphics[width=\linewidth]{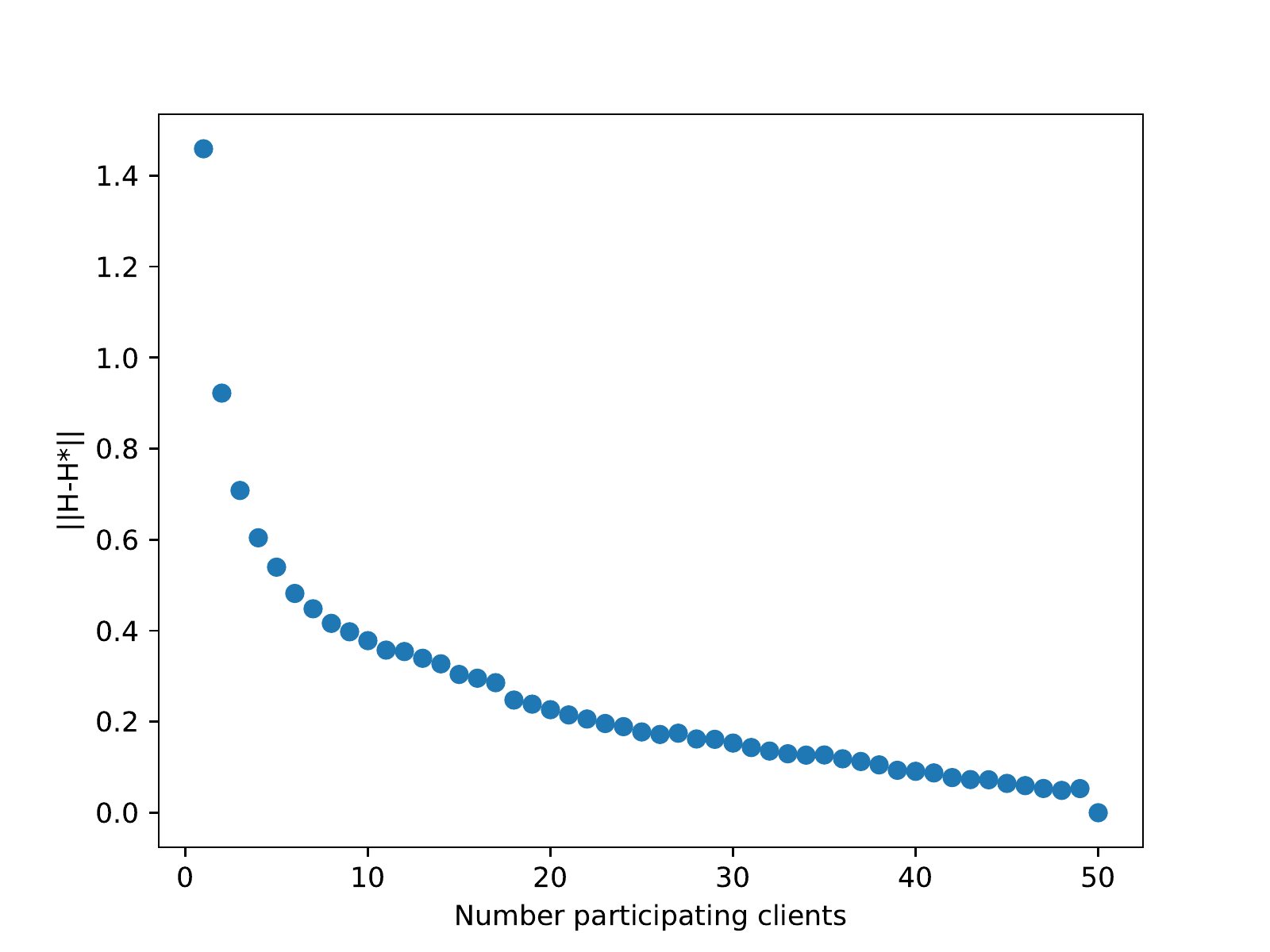}
        \caption{}
        \label{fig:HessianSimilarity}
     \end{subfigure}
    \caption{All Experiments except for (d) are on our synthetic dataset. All Experiments are in cross-device setting. (a) Variations of GIANT on i.i.d. synthetic data (b) Variation of LocalNewton on synthetic data (c) Comparison of methods using two procedural rounds of communication (d) Experiment on w8a where \textit{Local Newton with global line search} and Federated Averaging have the same budget of gradient evaluations. We use the average of gradient evaluations from \textit{Local Newton with global line search} although the cg method needs more iterations closer to the optimum. (e) Sampling another set of clients for the global line search to not "overfit" to the currently selected client does not improve performance except for \textit{Local Newton with global line search} (f) Quality of estimation of overall Hessian by increasing number of local Hessians on w8a. The norm using the identity matrix as $H^*$ as in Federated Averaging is circa 17. 5 of the 50 clients are participating in each round of our federated learning experiments.}
\end{figure*}

\end{document}